\documentclass{article}
\usepackage{ijcai24}

\usepackage{times}
\usepackage{soul}
\usepackage{url}
\usepackage[hidelinks]{hyperref}
\usepackage[utf8]{inputenc}
\usepackage[small]{caption}
\usepackage{graphicx}
\usepackage{amsmath}
\usepackage{amsthm}
\usepackage{booktabs}
\usepackage{algorithm}
\usepackage{algorithmic}
\usepackage[switch]{lineno}

\usepackage{comment}

\usepackage{amsfonts}
\usepackage{microtype}
\usepackage{multirow}
\usepackage[table]{xcolor}

\usepackage[table]{xcolor}
\usepackage{etoolbox}
\usepackage{pgf} 
\usepackage{pifont}
\newcommand{\xmark}{\textcolor[RGB]{200, 200, 200}{\ding{55}}}%
\definecolor{high}{HTML}{c2000a}  
\definecolor{mid}{HTML}{ffffff}  
\definecolor{low}{HTML}{0a9403}  
\newcommand*{\opacity}{80}
\newcommand*{\minval}{0.15}
\newcommand*{\midval}{1.0}
\newcommand*{\maxval}{1.85}

\newcommand{\gradient}[1]{
    \ifdimcomp{#1pt}{>}{\maxval pt}{#1}{
        \ifdimcomp{#1pt}{<}{\minval pt}{#1}{
            \ifdimcomp{#1pt}{<}{\midval pt}{
                \pgfmathparse{int(round(100*(#1/(\midval-\minval))-(\minval*(100/(\midval-\minval)))))}
                \xdef\tempa{\pgfmathresult}
                \cellcolor{mid!\tempa!low!\opacity} #1
            }{
                \pgfmathparse{int(round(100*(#1/(\maxval-\midval))-(\midval*(100/(\maxval-\midval)))))}
                \xdef\tempa{\pgfmathresult}
                \cellcolor{high!\tempa!mid!\opacity} #1
    }}}
}

\newtheorem{example}{Example}
\newtheorem{theorem}{Theorem}

\title{Neural Networks with Causal Graph Constraints: \\ A New Approach for Treatment Effects Estimation}

\author{
Roger Pros$^{1,2}$
\and
Jordi Vitrià$^1$
\affiliations
$^1$Departament de Matemàtica i Informàtica,
Universitat de Barcelona\\
$^2$Zenital\\
\emails
\{roger.pros, jordi.vitria\}@ub.edu,
}

\begin{document}

\maketitle

\begin{abstract}
In recent years, there has been a growing interest in using machine learning techniques for the estimation of treatment effects. Most of the best-performing methods rely on representation learning strategies that encourage shared behavior among potential outcomes to increase the precision of treatment effect estimates. In this paper we discuss and classify these models in terms of their algorithmic inductive biases and present a new model, NN-CGC, that considers additional information from the causal graph. NN-CGC tackles bias resulting from spurious variable interactions by implementing novel constraints on models, and it can be integrated with other representation learning methods. We test the effectiveness of our method using three different base models on common benchmarks. Our results indicate that our model constraints lead to significant improvements, achieving new state-of-the-art results in treatment effects estimation. We also show that our method is robust to imperfect causal graphs and that using partial causal information is preferable to ignoring it.
\end{abstract}

\section{Introduction}


Causal inference is a method used to understand whether and how a specific action (intervention or treatment) causes a change in a particular result or outcome. This process is crucial in many fields where understanding the cause-effect relationship is important; such as in medical trials, economic policy analysis, educational program evaluations, social science research, and artificial intelligence \cite{scholkopf2021toward}.


A key challenge in causal inference is that when we observe an outcome, we usually do not know what would have happened with an alternative intervention. This issue is known as the ``Fundamental Problem of Causal Inference" \cite{rubin1974estimating}. Due to this, direct application of data-based function approximation methods, such as machine learning or statistical analysis, is not feasible. As a result, various indirect methods have been devised to address this challenge.


In the case of observational data, that is, information collected without any deliberate intervention or controlled experiment, causal inference faces unique challenges and approaches. 

The first major challenge in conducting causal inference from observational data is the identifiability problem \cite{pearl2009causality}: establishing that the causal effect can be uniquely inferred from the data at hand. This requires ensuring that the data, model, and assumptions used are appropriate and robust enough to distinguish between correlation and causation. 

Once identifiability is established, the next step is to articulate the target causal effect using estimable quantities. Machine learning techniques have proven highly effective in calculating the types of expressions found in these estimable quantities, playing a significant role in advancing various causal inference methodologies \cite{causalml_survey}.

During recent years, there has been an increasing emphasis on developing techniques that incrementally adapt machine learning models to the task of causal effects estimation. This area, where representation learning and neural networks play a vital role \cite{tarnet,dragonnet,bcauss}, is the focus of our paper. In this work, we review and categorize these models based on their algorithmic inductive bias – the underlying assumptions a model relies on for making predictions or generalizations from its training data. We explore their relationship with various types of biases that need to be addressed for accurate treatment effect estimation.


Furthermore, we pinpoint and address a bias that is often overlooked in the literature on non-linear models: the occurrence of spurious interactions between variables within models. These interactions are not part of the underlying causal model and can create correlational shortcuts that alter the impact of causal relationships estimated by models, particularly in situations where data is limited.

The rest of this paper is organized as follows. In Section 2, we define the problem and review related work on causal effects estimation. In Section 3, we describe our proposed method and explain how it can be implemented in neural network models. In Section 4, we present the experiments we conducted to evaluate our method and compare it with other state-of-the-art estimators. In Section 5, we discuss the results and analysis of our experiments and suggest some possible directions for future research.

\subsection{Preliminary concepts and setup}

We want to estimate the causal effect of a treatment variable $T$ on an outcome variable $Y$, controlling for a set of covariates $X$. We assume that the causal relationships between these variables can be represented by a \textbf{causal graph} \cite{causal_graph}. A causal graph is a directed acyclic graph (DAG) where each node is a variable and each edge is a direct causal effect. We assume that the causal graph satisfies the causal edges assumptions, which means that all parents are causes of their children. We denote the causal graph by $G$, the set of parents of node $x_i$ by $Pa(x_i)$, the children by $Ch(x_i)$ and the ancestors by $An(x_i)$.

Beyond the causal graph, the underlying causal model can be modeled through a set of structural equations. A \textbf{Structural Causal Model}  (\textsc{scm}) \cite{pearl2009causality} consists of a set of endogenous variables $U$, a set of exogenous variables $V$ and a set of functions $\{f_i\}$ to generate each endogenous variable $x_i$ as a function of other variables. Every \textsc{scm} implies an associated causal graph. 

An {\bf interaction} occurs when the effect of one variable on the outcome depends on the value of another variable. However, models for treatment effect estimation can consider interactions that may be spurious, lacking a real presence in any causal mechanism within the data generation process. Hence, we define a \textbf{spurious interaction} as one that does not manifest in any of the equations $f_i$ within the \textsc{scm}. We refer to non-spurious interactions as \textbf{causally valid interactions}. 

In many cases, the complete \textsc{scm} may not be fully known, but there might be some expert knowledge about the nature of the interactions between variables. In such situations, the method proposed in this paper can still be utilized by incorporating this partial information.

In this work, we will focus on causal effects that can be identified using the \textbf{backdoor criterion} \cite{backdoor_criterion}. A set of variables $X$ satisfies the backdoor criterion relative to $T$ and $Y$ if: 1) $X$ blocks all backdoor paths from $T$ to $Y$, and 2) $X$ does not contain any descendants of $T$. 

\subsubsection{Adjustment sets and variable interactions}

Even though the backdoor criterion provides a complete identification of all {\bf valid adjustment sets} (sets of variables can be used to identify the causal effects, possibly including variables that are not strictly necessary), it does not tell us which is the {\bf optimal adjustment set} for practical estimation in finite-sample scenarios. 

Those variables in a valid adjustment set that are not strictly required for identification are called \textbf{neutral controls} \cite{good_controls}. These variables neither increase nor decrease the asymptotic bias of the estimation of treatment effects, but they can still have a huge impact in a finite-sample scenario and can improve the performance of estimators. The optimal adjustment set usually contains neutral controls. 

Given a valid adjustment set, its associated spurious interactions are the potential interactions between variables included in the set that do not appear in any of the of the equations of the  \textsc{scm} associated with the data.

In this work, we assume all our datasets are composed of valid adjustment sets, i.e. the backdoor criterion can be satisfied. There are advanced methods that, given an  \textsc{scm}, allow to define {\bf efficient adjustment sets}  (adjustment sets that yield the least squares estimator with the smallest asymptotic variance) \cite{smucler}, \cite{henckel2022graphical}. Although this scenario is not addressed in our work, it's important to note that these sets may also be prone to spurious interaction bias. Consequently, the method outlined in this paper is also applicable to them.

\section{Problem definition and related work} 

We are interested in estimating the $f_Y(T, X)$ function that explains the causal effect of a given treatment $T$ on a target variable $Y$ considering a valid adjustment set $X$. In this paper, we name this the \textit{causal effects function}.

In the binary case, let us define $Y(1)$ as the outcome in the presence of treatment and $Y(0)$ as the outcome in its absence. For a given individual $i$, the causal effects function, or Individual Treatment Effect function (\textsc{ite}), is expressed as $ITE:= f_Y(T, X) = f_Y(X) = Y_i(1) - Y_i(0).$ We are also interested in the Average Treatment Effect (\textsc{ate}) and the Conditional Average Treatment Effect (\textsc{cate}): $ATE := \mathbb E[Y_i(1) - Y_i(0)],$
$CATE := \mathbb E[Y_i(1) - Y_i(0) | X=x].$

In order to be able to estimate these quantities, we need to obtain a statistical estimand that depends only on observed data using the process of identification.  The identified estimand for the binary case using the backdoor criterion is:
\begin{equation}
    \mathbb E_X[\mathbb E[Y|T=1, X] - \mathbb E[Y|T=0, X]]
    \label{Eq1}    
\end{equation}

This estimand is apparently similar to the common estimand used in prediction tasks $\mathbb E[Y|T, X]$, but model selection processes and inductive biases that work for one do not necessarily apply to the other \cite{selectcausal}. Consequently, having an identified expression might not be enough to calculate accurate causal effects. The estimand that has been identified might be prone to issues that could make it difficult to estimate and hence render any non-asymptotic estimation impossible \cite{estimability}.

The step after causal identification is causal estimation: in this step, all available data is used to compute an association that reflects the causal effect if all assumptions are met.

\subsection{Causal Estimation and Inductive biases for the causal inference task}

The use of algorithmic inductive biases that encode prior information about the data in causal estimation methods has been one of the most successful strategies to build better causal inference models \cite{kunzel2019metalearners}. To understand these inductive biases, we need to identify the sources of error when computing causal effects. These sources of error can be categorized into two main categories: identification bias and estimation bias \cite{chatton2023causal}.
 
In practical situations, it is not always possible to test the assumptions that establish a statistical estimand as a causal estimand. Certain assumptions, such as the absence of confounders, cannot be tested. Below, we present several biases that arise from the potential violation of the assumptions, known as \textbf{identification bias}:

\begin{table*}[h]
  \centering
  \resizebox{1\textwidth}{!}{
  \begin{tabular}{c|c|c|c|c|c|c|c|c}
    \toprule
      \multirow{2}{*}{\textsc{Method}}  & \multicolumn{4}{c|}{\textsc{Identification Bias}} & \multicolumn{4}{c}{\textsc{Estimation Bias}}  \\
    & Exchangeability & Positivity & No interference & Consistency & Data eficiency & Treatment relevance & Propensity  & Spurious interactions  \\
    \midrule
    \rowcolor{gray!10} Linear Models \cite{thoresen2019spurious} & \xmark & \xmark & \xmark  & \xmark & \checkmark & \xmark & \xmark &  \checkmark \\ 
    S-learner \cite{kunzel2019metalearners} & \xmark &  \xmark&  \xmark &\xmark  & \checkmark & \xmark & \xmark & \xmark \\
    \rowcolor{gray!10} T-learner \cite{kunzel2019metalearners}  & \xmark &  \xmark & \xmark  & \xmark & \xmark & \checkmark &  \xmark & \xmark \\
    X-learner \cite{kunzel2019metalearners} & \xmark & \xmark & \xmark  &\xmark  & \checkmark & \checkmark & \xmark  & \xmark \\
    \rowcolor{gray!10} BLearner \cite{blearner}  &  \checkmark &\xmark  & \xmark  & \xmark & \checkmark & \checkmark & \checkmark & \xmark \\ 
    TARNet\cite{tarnet} &  \xmark & \xmark & \xmark & \xmark  & \checkmark &  \checkmark & \xmark &  \Large{+} \textsc{cgc} \Large{$\longrightarrow$} \phantom{.}  \checkmark   \\
    \rowcolor{gray!10}CFRNet \cite{tarnet}  & \xmark  & \xmark &  \xmark &   \xmark & \checkmark &  \checkmark &  \checkmark &    \Large{+} \textsc{cgc} \Large{$\dashrightarrow$} \phantom{.}  \checkmark ?  \\ 
    Dragonnet \cite{dragonnet}  &  \xmark & \xmark  & \xmark  &\xmark  & \checkmark &  \checkmark &  \checkmark & \Large{+} \textsc{cgc} $\longrightarrow$ \phantom{.}  \checkmark \\ 
    \rowcolor{gray!10}BCAUSS \cite{bcauss}  & \xmark &  \checkmark & \xmark  & \xmark & \checkmark &  \checkmark &  \checkmark &\Large{+} \textsc{cgc} $\longrightarrow$ \phantom{.} \checkmark  \\
    HINITE \cite{lin2023estimating} & \xmark & \xmark &  \checkmark & \xmark  & \checkmark &  \checkmark &  \checkmark &  \Large{+} \textsc{cgc} \Large{$\dashrightarrow$} \phantom{.}  \checkmark ? \\
    \bottomrule
  \end{tabular}
  }
  \caption{Comparison of some relevant causal inference methods in terms of the biases they address. Checkmarks indicate that the method has a mechanism to address the bias indicated in the column. Methods included in the table have mechanisms that are easy to associate with a source of error. In this paper, we present CGC, a method that can be applied on top of other representation-based learners to address the spurious interactions bias. With a solid arrow, we represent the methods evaluated in this paper. With a dashed arrow, other representation-based methods where CGC could be applied.}
   \label{table:comp}
\end{table*}

    \begin{itemize}
        \item Exchangeability bias: Bias derived from the existence of hidden confounders that impact the identification process.
        \item Positivity bias: This assumption can be violated if we do not collect any data for a subpopulation. 
        \item Consistency bias: Bias derived from the violation of the consistency assumption, which states that if the treatment is $T$, then the observed outcome $Y$ is the potential outcome under treatment $T$ \cite{neal2020introduction}. 
        \item Non-interference bias: Bias derived from the effect of one individual treatment on another individual. The existence of this bias is particularly common in network data \cite{lin2023estimating}.

    \end{itemize}

We can also consider additional inductive biases related to the finite nature of the data. These are called \textbf{estimation bias}:
    \begin{itemize}
        \item Propensity bias: A causal model must generalize from behavior under one set of conditions to behavior under another set \cite{pearl2009causality}. This bias appears when the performance of the model differs between distinct treatments as a consequence of a difference in the distribution of the data.
        \item Relevance of treatment: Bias that appears especially in high-dimensional settings. Models tend to neglect the role of the treatment and its effect is expressed through correlated features.
        \item Functional form: Any task performance can be affected by the functional form in which it is modeled. Linear tasks are better modeled by linear models, and monotonic tasks by monotonic models.
        \item Variable interactions: Spurious variable interactions do not explain any causal relationship between the variables, but a model might use them as a correlational shortcut, modifying the effect of causal relationships. 
        \item Data efficiency: A model might not make an efficient use of the information present in the data.
        \item Missing data: Bias derived from the lack of availability of some data.
        \item Measurement error on some variables.
    \end{itemize}

In Table \ref{table:comp} some of the most relevant causal inference methods are considered along with the biases they address. It shows which methods can correct or mitigate each type of bias. Using data efficiently and ensuring treatment relevance are among the most critical sources of error and have been addressed by most models. In the table, we can also see that one of the most interesting properties of non-linear models is that they can stack multiple bias correction strategies to address several sources of bias at the same time.

\subsection{State of the Art causal inference methods}

Two of the most straightforward approaches for estimating the estimand represented by Eq.(\ref{Eq1}) are the S-Learner and the T-Learner \cite{kunzel2019metalearners}. S-learner is a method that trains a single machine learning model with the treatment variable as one of the features, along with other covariates. This model is then used to predict outcomes under both scenarios - with treatment and without treatment. The main disadvantage of S-learner is that it can bias the estimate of the treatment effect towards zero, especially when the feature space is large. Unlike the S-learner, the T-learner splits the data by treatment and fits a model for each group. This  addresses the issue of failing to recognize a weak treatment variable, yet it may still be prone to data efficiency bias.

S-learner and T-learner are among the simplest causal effects estimators; however, apart from these approaches, more sophisticated methods have also been developed.

Most of the best performing methods in Table \ref{table:comp} rely on the concept of \textbf{representation learning} \cite{bengio2013representation}. They use neural networks with an architecture that can be understood in two parts: pre-representation and post-representation. The first creates a representation from the input data and the second takes over from the representation to the desired output. One reason why explicitly dealing with representations is interesting is because they can be convenient to express many general priors about the world around us. These priors can be used to incorporate inductive biases to improve the causal inference task. Recent improvements in causal effect prediction methods focus primarily on modifications of the post-representation part: either by modifying the architecture of the output layers of the network or by using specific loss functions that carry inductive biases.

For example, TARNet \cite{tarnet}  branches off into two separate heads for each treatment group to prevent the model from disregarding $T$. On top of that, Dragonnet \cite{dragonnet} is adds a third head to the TARNet architecture to predict propensity $g(X) = P(T=1|X)$. This approach reduces the bias caused by having different modeling performances for each value of the treatment, although it decreases the predictive accuracy of the model. BCAUSS \cite{bcauss} uses the same architecture as the Dragonnet, but modifies the loss function so that it is more robust to positivity violations. In this approach, multiple bias correction methods are stacked to obtain state-of-the-art results. Other representation learning approaches encode other inductive biases to deal with different scenarios. CFRNet \cite{tarnet} and BNN \cite{bnn} create a balanced representation of the data for both treated and untreated observations. Similarly, importance sampling weights can be added to the CFRNet to alleviate the problem of selection bias \cite{hassanpour2019counterfactual}. HINITE \cite{lin2023estimating} addresses no interference assumption violations, and in \cite{snet} the authors create a representation that encodes prior information about the structure of causal effects.

Other works have suggested inductive biases that are challenging to link with a particular source of error. For example, CEVAE \cite{cevae} uses variational autoencoders (VAE) to estimate a latent-variable model that simultaneously discovers hidden confounders and infers how they affect treatment and outcome. Another example is GANITE \cite{yoon2018ganite}, which uses generative adversarial networks (GAN) to estimate treatment effects. This approach attempts to capture the uncertainty in the counterfactual distributions and is defined for any number of treatments.

\subsection{Motivation} 

The source of bias derived from variable interactions has been studied in the context of linear models \cite{thoresen2019spurious}, \cite{spurious_linear}, where the inclusion of each interaction has to be evaluated. However, the discussion has not extended to non-linear models like neural networks, despite these models frequently rely on interactions between all variables.

In this paper we propose Neural Networks with Causal Graph Constraints (NN-CGC), a method to address the variable interactions bias using causal information like the one that might be present in the causal graph and apply it to neural network models.

\section{A new Inductive Bias for Estimating Causal Effects}


We propose  a comprehensive methodology for incorporating an inductive bias that mitigates spurious interactions in neural network models. This approach can utilize information such as expert insights about the connections between specific variables, partial or complete knowledge of the causal graph, or a blend of both.

\subsection{Constraining the learned distribution}
Our main assumption is that the learned distribution has to be as close as possible to the distribution defined by the underlying causal model. To do so, we remove some of the cases that do not satisfy the conditional independence structure encoded by the causal graph. Specifically, we model the distribution of $Y$ as a function of functions of variables that are causally related to each other.

For any node $x_i \in Pa(Y)$, we define the following set of variables: 
$G_{x_i} := (\{x_i\} \cup An(x_i) ) \setminus \{T\}$. 

Then, we propose to restrict the learned model $f_Y(X)$ to this family of functions: $$f_Y(X) \sim  f_Y\Big( f\big(Pa(Y) \setminus \{T\} \big), f\big(G_{x_1}\big), .., f\big(G_{x_n}\big)\Big).$$
The variable $T$ is excluded from all the groups because the learners implemented in this paper do not use $T$ as an input but rather as part of the structure.

\begin{figure}[h]
    \centering
    \includegraphics[scale = 0.40]{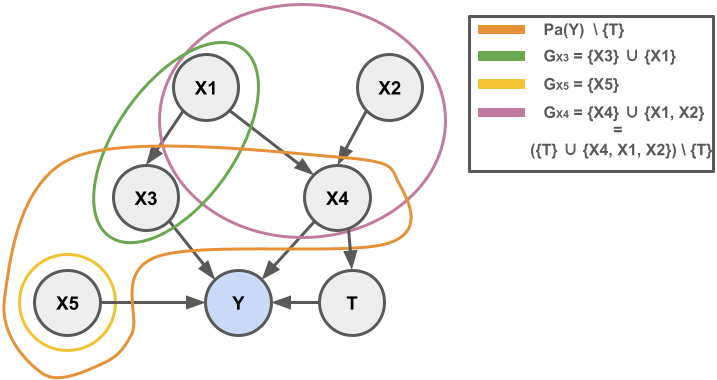}
    \caption{Illustrative example of groups of variables allowed to interact between each other. In this setting, the allowed interactions are: (X1, X2, X4), (X3, X1), (X5, X3, X4). In this specific case, $G_{x4}$ is equal to $G_T$.}
    \label{fig:dag}
\end{figure}

 Using this method, all valid interactions are included, while all discarded interactions are of spurious nature. Note that we do not use causal paths that go from root nodes to the target as groups, as these would prevent some valid interactions.
See an example in Figure \ref{fig:dag}.

In the case of not having access to the causal graph, we can be define $G_{x_i}$ by using expert knowledge about existing spurious interactions or by considering causal discovery methods. This approach can help limit our model's reliance on at least some of the spurious interactions.

\subsubsection{Groups of variables and Independent Causal Mechanisms}
The causal generative process of a system variables
is composed of autonomous modules that do not inform
or influence each other, called Independent Causal Mechanisms (ICM) \cite{scholkopf2021toward}. The groups defined in Section 3.1 contain all the information for modeling the ICMs for the target variable $Y$ and for each of its parents $Pa(Y)$ given the adjustment set $X$. Several studies, such as those by \cite{xia2021causal} and \cite{parafita2022estimand}, have investigated the process of modeling all Independent Causal Mechanisms (ICMs) to derive a Structural Causal Model (SCM). This comprehensive method carries its own challenges and advantages, making it a suitable option for certain scenarios. The methodology presented in this paper finds a balance between fully connected modeling and SCM-based modeling.


Following this objective, in the next section, we detail how this approach can be implemented using neural networks and how it can be combined with other neural network-based learners.

\begin{figure}[h]
\centering
\includegraphics[scale = 0.23]{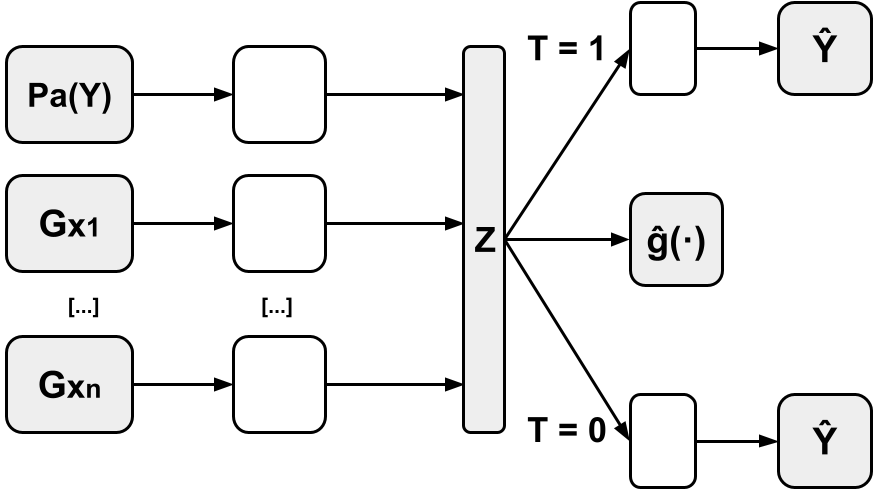}
\caption{Model architecture when applying CGC to the Dragonnet. The post-representation part remains identical but the pre-representation layers are divided according to the groups of variables.}
\label{fig:architecture}
\end{figure}


\subsection{Implementation and Neural Network Architecture}

Similarly to other methods related to representation learning ideas, our technique can be broken down into two parts. Initially, we define the inputs and the layers that lead to the representation. Subsequently, we analyze the representation layer and the output layers of the network. A pseudocode for the implementation can be found in Algorithm \ref{alg:cgc}.

\subsubsection{Pre-representation: Variable groups}

The input of the network is divided into different groups of variables, as shown in Figure \ref{fig:architecture}. We create a set of layers for each variable group $G_{x_i}$ and $Pa(Y)$, that do not interact with each other. Each set of layers receives only the variables in the group as input. Note that a single variable can be part of more than one group. Each set of layers must be able to model $P(Y | G_{x_i})$ or $P(Y | Pa(Y))$. The output of all these layers is concatenated and used as input of the representation layer $Z$.

\subsubsection{Post-representation: Linear layer and output heads}

The representation layer $Z$ is connected to the head of the network through a linear activation function.  This ensures that the representation encoded in $Z$, which is used as input for the rest of the network, is free from spurious interactions.


Most state of the art methods modify the output of the network by changing the architecture after the representation layer, i.e. considering distinct heads or loss functions. In contrast, our methodology takes place in the architecture before the representation layer. This allows for a combination of both architectures, using our methodology combined with any of the output heads. In this paper we implemented the TARNet \cite{tarnet}, Dragonnet \cite{dragonnet} and BCAUSS \cite{bcauss} heads, but other representation-based learners can be connected to the network. 

\section{Empirical testing}
We are interested in determining whether models that are constrained using the proposed technique perform better than their unconstrained counterparts. Since the method can be applied to most representation-based learners, we test it using some of the best performing learners:TARNet, Dragonnet, and BCAUSS. Counting constrained and unconstrained versions, this gives us a total of six models to experiment with.

\begin{algorithm}[tb]
    \caption{Adding CGC procedure}
    \label{alg:cgc}
    \textbf{Input}: Train and test sets $(D_{tr}, D_{te}) \sim D$ with a treatment variable $T$, a target $Y$ and a valid adjustment set $X$. \\
    \textbf{Parameter}: A representation-based $f_Y(T, X)$ estimator \textit{NN}.\\
    \begin{algorithmic}[1] 
        \vspace{-0.3cm}
        \STATE Causal Knowledge Specification: Build or discover the associated causal graph using the data in $D_{tr}$. If this is not feasible, list all forbidden variable interactions by using expert knowledge.
        \vspace{0.2cm}
        \STATE Groups: Create the groups of variables $G_{x_i} := (\{x_i\} \cup An(x_i) ) \setminus \{T\} \phantom{.} \forall x_i \in Pa(Y)$ and $Pa(Y)\setminus \{T\}.$ If the causal graph is not available, form all compatible variable subsets by excluding forbidden interactions.
        \vspace{0.2cm}
        \STATE Architecture pre-representation: Create an independent set of layers for each group, all connected to the representation layer $Z$.
        \vspace{0.2cm}
        \STATE Architecture post-representation: Use the head associated to \textit{NN} from the representation layer onward.
        \vspace{0.2cm}
        \STATE Model: Run the causal inference workflow associated to \textit{NN} using $D_{tr}$ and $D_{te}$.     
    \end{algorithmic}
\end{algorithm}

\subsection{Experimental setup}

Evaluating causal inference methods is hard because we usually have no access to counterfactual data. To overcome this problem, one of the common approaches is to use semi-synthetic data. We test the models on two of the most common benchmarks in causal inference: \textsc{IHDP} and \textsc{Jobs}. These benchmarks have already been used by some of the best performing methods and thus are specially relevant for the comparison of constrained and unconstrained models.

Since these benchmarks might present some problems \cite{curthbenchmark}, we also test the models on a range of synthetic benchmarks that represent distinct situations, details in the synthetic experiments section. All the code for reproducing the experiments can be found \href{https://anonymous.4open.science/r/NN-CGC-FF9E/}{here} - https://anonymous.4open.science/r/NN-CGC-FF9E/.

\subsubsection{Causal Graphs}
All causal graphs in these experiments are discovered using the \textsc{IcaLinGam} algorithm \cite{icalingam} implemented in \cite{zhang2021gcastle}. The output of this algorithm usually involves only a subset of the variables. Since we are interested in the effect of using correct information rather than the quality of the discovery, we group the remaining variables into a single group of variables. Errors in the discovery might still translate to loss of performance of the constrained models (which can be seen in the higher-variance scenarios of the synthetic benchmarks). The choice of the discovery algorithm is out of the scope of this work, and \textsc{IcaLinGam} has been selected for its ability to converge under distinct scenarios. Given that the adjustment set is considered valid, it is assumed that any nodes associated with the target are in the parents of $Y$, as this would otherwise go against the backdoor criterion.

\subsubsection{Hyperparameters}

We use the same training hyperparameters defined in  \cite{dragonnet} and  \cite{bcauss}, as they show to achieve state of the art results. These include a hidden layer size of 200 for the shared representation layers and 100 for the conditional outcome layers, train using stochastic gradient descent with learning rate $10^{-5}$ and momentum $0.9$, and batch size equal to 64 for the TARNet and Dragonnet and equal to the train-set length for BCAUSS. Details can be found in the code implementation and in appendix B.

In the constrained architectures, the hyperparameters for each set of layers are kept the same as those in the unconstrained architectures. However, the activation function of the representation layer is modified to a linear function.

\subsubsection{Evaluation metrics}

When counterfactual data is accessible, we use standard metrics such as the error in Average Treatment Effects -\textsc{ate}- and the Precision in Estimation of Heterogeneous Effects -\textsc{pehe}.


For the Jobs dataset, no counterfactuals are available, so we use the experimental nature of the data to compute the error of the average treatment effects on the treated \textsc{att}.



\begin{table}[t]
\centering
\resizebox{0.40\textwidth}{!}{
\begin{tabular}{|c|c|c|c|c|}
\toprule
\textsc{scenario}      & \textsc{noise}  &   \scalebox{0.8}{\textsc{TARNet}}   &   \scalebox{0.8}{\textsc{Dragonnet}}  &  \scalebox{0.8}{\textsc{BCAUSS}} \\
\midrule
\multirow{4}{*}{A}  & 1/2  & \gradient{0.76}  & \gradient{0.76}  & \gradient{0.93} \\ \cline{2-5}
& 1  & \gradient{0.91}  & \gradient{0.92}  & \gradient{1.02} \\  \cline{2-5}
& 2  & \gradient{1.07}  & \gradient{1.04}  & \gradient{1.06} \\  \cline{2-5}
& 4  & \gradient{1.08}  & \gradient{1.11}  & \gradient{1.02} \\
\midrule
\multirow{4}{*}{B}  & 1/2  & \gradient{0.73}  & \gradient{0.72}  & \gradient{0.63} \\ \cline{2-5}
& 1  & \gradient{0.92}  & \gradient{0.94}  & \gradient{0.78} \\  \cline{2-5}
& 2  & \gradient{1.02}  & \gradient{1.06}  & \gradient{0.94} \\  \cline{2-5}
& 4  & \gradient{1.00}  & \gradient{0.99}  & \gradient{0.96} \\
\midrule
\multirow{4}{*}{C}  & 1/2  & \gradient{0.51}  & \gradient{0.51}  & \gradient{0.36} \\ \cline{2-5}
& 1  & \gradient{0.71}  & \gradient{0.76}  & \gradient{0.7} \\  \cline{2-5}
& 2  & \gradient{1.00}  & \gradient{1.}  & \gradient{1.09} \\  \cline{2-5}
& 4  & \gradient{1.12}  & \gradient{1.13}  & \gradient{1.18} \\
\midrule
\multirow{4}{*}{D}  & 1/2  & \gradient{0.5}  & \gradient{0.51}  & \gradient{0.38} \\ \cline{2-5}
& 1  & \gradient{0.76}  & \gradient{0.76}  & \gradient{0.72} \\  \cline{2-5}
& 2  & \gradient{0.92}  & \gradient{0.92}  & \gradient{0.93} \\  \cline{2-5}
& 4  & \gradient{1.00}  & \gradient{1.00}  & \gradient{1.01} \\
\bottomrule
\end{tabular}
}
\caption {Table for PEHE Comparison. The reported value is the ratio between the constrained and the unconstrained models. Values $<1$ mean the constrained version has a lower error than the unconstrained counterpart.} 
\label{tab:PEHEsyn} 
\end{table}

\begin{table}[t]
\centering
\resizebox{0.40\textwidth}{!}{
\begin{tabular}{|c|c|c|c|c|}
\toprule
\textsc{scenario}      & \textsc{noise}  &   \scalebox{0.8}{\textsc{TARNet}}   &   \scalebox{0.8}{\textsc{Dragonnet}}  &  \scalebox{0.8}{\textsc{BCAUSS}} \\
\midrule
\multirow{4}{*}{A}  & 1/2  &  \gradient{0.73 }  &  \gradient{ 0.74 }  &  \gradient{ 0.93} \\ \cline{2-5}
& 1  &  \gradient{ 0.83 }  &  \gradient{ 0.81  }  &  \gradient{1.02} \\  \cline{2-5}
& 2  &  \gradient{0.96 }  &  \gradient{ 0.92  }  &  \gradient{ 1.07} \\  \cline{2-5}
& 4  &  \gradient{1.01 }  &  \gradient{ 1.04}  &  \gradient{ 1.02} \\
\midrule
\hline
\multirow{4}{*}{B}  & 1/2  &  \gradient{0.87  }  &  \gradient{ 0.88 }  &  \gradient{ 0.69 } \\ \cline{2-5}
& 1 &  \gradient{0.96  }  &  \gradient{ 0.98  }  &  \gradient{0.87} \\  \cline{2-5}
& 2  &  \gradient{0.95  }  &  \gradient{0.99  }  &  \gradient{ 0.98} \\  \cline{2-5}
& 4&  \gradient{ 0.91 }  &  \gradient{0.93  }  &  \gradient{ 0.95} \\
\midrule
\multirow{4}{*}{C}  & 1/2 &  \gradient{.58 }  &  \gradient{ 0.58}  &  \gradient{ 0.17} \\ \cline{2-5}
& 1  &  \gradient{0.54 }  &  \gradient{0.64 }  &  \gradient{0.31} \\  \cline{2-5}
& 2  &  \gradient{0.70 }  &  \gradient{0.72 }  &  \gradient{0.54 }\\  \cline{2-5}
& 4  &  \gradient{0.87 }  &  \gradient{0.91 }  &  \gradient{ 0.79 }\\
\midrule
\multirow{4}{*}{D}  & 1/2  &  \gradient{ 0.85 }  &  \gradient{0.92}  &  \gradient{0.7}  \\ \cline{2-5}
& 1  &  \gradient{0.98}  &  \gradient{1.01}  &  \gradient{ 0.88} \\  \cline{2-5}
& 2  &  \gradient{1.01}  &  \gradient{0.99 }  &  \gradient{0.94} \\  \cline{2-5}
& 4  &  \gradient{1.02}  &  \gradient{1.03}  &  \gradient{1.07} \\
\bottomrule
\end{tabular}
}
\caption {Table for ATE Comparison. The reported value is the ratio between the constrained and the unconstrained models. Values $<1$ mean the constrained version has a lower error than the unconstrained counterpart.} 
\label{tab:ATEsyn} 
\end{table}

\subsection{Experiments}
\subsubsection{Synthetic experiments}

We use a set of synthetic experiments defined in \cite{rlearner} and implemented in \cite{chen2020causalml}. These experiments are composed of four scenarios: scenario A is a complex outcome regression model with an easy treatment effect, scenario B is a randomized controlled trial, scenario C has an easy propensity score with a difficult control outcome, and scenario D has an unrelated treatment arm and control arm. We run each scenario using the settings in the original paper: number of samples $n \in \{500, 1000\}$, number of variables $d \in \{6, 12\}$, and noise level $\sigma \in \{0.5, 1, 2, 4\}$. We generate an equal number of test samples as the number of train samples. Each configuration is repeated 100 times, making a total of 6400 runs. We average for the number of samples and number of variables for controlling the reporting size and show the results at the scenario-noise level because it contains the most information. Detailed results can be found in appendix A.

\begin{table}[h]
\label{tab:title} 
\centering
\resizebox{0.48\textwidth}{!}{
\begin{tabular}{c|c|c|c|c|c|c|c|c}
\toprule
  \multirow{2}{*}{\textsc{Model}}  & \multicolumn{4}{c|}{$\sqrt{\epsilon_{PEHE}}$} & \multicolumn{4}{c}{$\epsilon_{ATE}$}  \\
    & A & B & C & D & A & B & C & D  \\

\midrule 
TARNet & .363 & .350 & .456 & .564 & .285 & .197 & .316 & .158 \\
Dragonnet & .359 & .349 & .446 & .568 & .283 & .196 & .306 & .160 \\
BCAUSS & .413 & .432 & .424 & .681 & .346 & .212 & .405 & .163 \\

\midrule
TARNet + CGC & .350 & \cellcolor{gray!25} .334 & .411 & \cellcolor{gray!25} .475 & .252 & \cellcolor{gray!25} .182 & .232 & .158 \\
Dragonnet + CGC & \cellcolor{gray!25} .347 & .335 & .408 & .478 & \cellcolor{gray!25} .250 & .185 & .238 & .161 \\
BCAUSS + CGC & .417 & .366 & \cellcolor{gray!25} .378 & .523 & .348 & .195 & \cellcolor{gray!25} .207 & \cellcolor{gray!25} .157 \\

\bottomrule
\end{tabular}
}
\caption {Test metrics on the synthetic benchmark averaged to the scenario level. Lower is better.} 
\label{tab:abssyn} 
\end{table}
\paragraph{Synthetic experiments results.}
The results of the synthetic experiments are shown in the metric values Table \ref{tab:abssyn} and in the comparison Table \ref{tab:ATEsyn} and Table \ref{tab:PEHEsyn}, where each value is the ratio between the \textsc{pehe} and \textsc{ate}, respectively, of the constrained and unconstrained models. A value less than 1 indicates that the constrained model has a lower metric value than the unconstrained model and vice versa. We observe that constrained models generally perform better than their unconstrained counterparts. Among the models with CGC, Dragonnet + CGC achieves the lowest \textsc{pehe} values in three out of four scenarios, while BCAUSS + CGC achieves the lowest \textsc{ate} values in two out of four scenarios. TARNet + CGC performs slightly worse than the other two models with CGC but still better than the models without CGC. In settings with higher variance, with a noise value 4 for the \textsc{ate}s and values 2 and 4 for the \textsc{pehe}s, the unconstrained models performed similar or better than the constrained models. This suggests that in high-noise environments the discerning between spurious and valid interactions is more difficult, and constraining the model can be discarding causally valid information.

\vspace{-1mm}
\subsubsection{Semisynthetic experiments: IHDP}
The Infant Health and Development Program (IHDP) is a study that used randomization to examine the effects of home visits by specialist doctors on the cognitive test scores of premature infants. The dataset\footnote{\label{fnote} The dataset can be accessed at {\tt www.fredjo.com}.} was first used by \cite{IHDP} to evaluate algorithms for estimating treatment effects. In order to create an observational dataset, non-random subsets of treated individuals were removed, resulting in selection bias. The outcomes in the dataset were generated using the original covariates and treatments. The dataset consists of 747 subjects and 25 variables. Following the recent literature, we used the simulated outcome implemented in the NPCI package \cite{dorie2016npci}, which is composed of 1000 repetitions of the experiment. We averaged our results over 1000 train/validation/test splits with ratios 70/20/10.


\vspace{-2mm}
\begin{table}[h]
\label{tab:title} 
\centering 
\resizebox{0.5\textwidth}{!}{
\begin{tabular}{c|c|c|c|c}
\toprule
  \multirow{2}{*}{\textsc{Model}}  & \multicolumn{2}{c|}{$\sqrt{\epsilon_{PEHE}}$} & \multicolumn{2}{c}{$\epsilon_{ATE}$}  \\
    & Value & Comp & Value & Comp  \\  
\midrule

BNN \cite{bnn}   & 2.2 $\pm$ .1 &  -  & .37 $\pm$ .03 &   -  \\
CFRW \cite{tarnet}   & .71 $\pm$ .0  &  -  & .25 $\pm$ .01 &   -  \\
CEVAEs  \cite{cevae} & 2.7 $\pm$ .1  &  -  & .34 $\pm$ .01 &   -  \\
GANITE \cite{yoon2018ganite}  & 1.9 $\pm$ .4  &  -  & .43 $\pm$ .05 &   -  \\
\midrule

TARNet + CGC   & .960 $\pm$ .032  &   \multirow{2}{*}{0.78}   & .148 $\pm$ .004 &   \multirow{2}{*}{1.04}   \\
TARNet                & 1.23 $\pm$ .046      &  & .142 $\pm$ .004  &  \\

\midrule
Dragonnet + CGC   & .974 $\pm$ .034 &  \multirow{2}{*}{0.797}   & .138 $\pm$ .003 &  \multirow{2}{*}{.968} \\
Dragonnet    & 1.22 $\pm$ .047 &  & .143 $\pm$ .004 &  \\
\midrule
BCAUSS + CGC & \cellcolor{gray!25} .663 $\pm$ .022  &   \multirow{2}{*}{.805}     &  \cellcolor{gray!25} .100 $\pm$ .003 & \multirow{2}{*}{.929} \\
BCAUSS      & .824 $\pm$ .023 &  & .107 $\pm$ .003 &  \\
\bottomrule
\end{tabular}
}
\caption {Train metrics on the IHDP benchmark. The ``Comp" value is the ratio between the constrained and the unconstrained model.} 
\end{table}

\vspace{-6mm}
\begin{table}[h]
\label{tab:title} 
\centering
\resizebox{0.5\textwidth}{!}{
\begin{tabular}{c|c|c|c|c}
\toprule
  \multirow{2}{*}{\textsc{Model}}  & \multicolumn{2}{c|}{$\sqrt{\epsilon_{PEHE}}$} & \multicolumn{2}{c}{$\epsilon_{ATE}$}  \\
    & Value & Comp & Value & Comp  \\

\midrule

BNN \cite{bnn}  & 2.1 $\pm$ .1 &  -  & .42 $\pm$ .03 &   -  \\
CFRW \cite{tarnet}  & .76 $\pm$ .0  &  -  & .27 $\pm$ .01 &   -  \\
CEVAEs  \cite{cevae} & 2.6 $\pm$ .1  &  -  & .46 $\pm$ .02 &   -  \\
GANITE  \cite{yoon2018ganite} & 2.4 $\pm$ .4  &  -  & .49 $\pm$ .05 &   -  \\
\midrule

TARNet + CGC   & 1.004 $\pm$ .039  &   \multirow{2}{*}{0.783}   & .180 $\pm$ .006 &   \multirow{2}{*}{0.884}   \\
TARNet                & 1.283 $\pm$ .052      &  & .204 $\pm$ .008  &  \\

\midrule
Dragonnet + CGC   & 1.027 $\pm$ .045 &  \multirow{2}{*}{0.796}   & .190 $\pm$ .009 &  \multirow{2}{*}{0.911} \\
Dragonnet    & 1.290 $\pm$ .057 &  & .209 $\pm$ .010 &  \\
\midrule
BCAUSS + CGC & \cellcolor{gray!25} .741 $\pm$ .030 &   \multirow{2}{*}{0.771}     &  \cellcolor{gray!25} .132 $\pm$ .005 & \multirow{2}{*}{0.862} \\
BCAUSS      & .962 $\pm$ .035 &  & .153 $\pm$ .006 &  \\
\bottomrule
\end{tabular}
}
\caption {Test metrics on the IHDP benchmark. The ``Comp" value is the ratio between the constrained and the unconstrained model.} 
\end{table}

\vspace{-3mm}
\paragraph{IHDP results.}
The constrained version of BCAUSS yields the best results in both \textsc{ate} and \textsc{pehe} metrics, with the unconstrained version of BCAUSS coming in second. In all cases, constrained models outperform unconstrained ones, obtaining lower errors and a similar or smaller variability in the results.

The \textsc{pehe} and \textsc{ate} metric values of the unconstrained models (TARNet, Dragonnet, and BCAUSS) were higher than those of the constrained models on both the train and test sets, indicating that the constraints reduced the estimation error and increased the generalization ability of the models.

\subsubsection{Real data experiments: Jobs} The LaLonde Jobs dataset\footref{fnote} \cite{lalonde1986evaluating} assesses the impact of job training as a treatment on income and employment status after training. We use the feature set defined in \cite{lalondefeatureset}. Following \cite{tarnet}, we combined the LaLonde experimental sample (297 treated, 425 control) with the PSID comparison group (2490 control). We averaged over 10 train/validation/test splits with a ratio of 62/18/20. Due to the small size of the benchmark we repeat the process 10 times and report the average of the results. For most of the runs, the discovery algorithm was unable to find a causal graph, and a default was used. The default causal graph was obtained as the most common causal graph discovered across 100 runs.
    
\begin{table}[h]
\label{tab:title} 
\centering
\resizebox{0.5\textwidth}{!}{
\begin{tabular}{c|c|c|c|c}
\toprule
\multirow{2}{*}{\textsc{Model}}  & \multicolumn{2}{c|}{$\epsilon_{ATT}^{tr}$} & \multicolumn{2}{c}{$\epsilon_{ATT}^{te}$}  \\
    & Value & Comp & Value & Comp   \\
\midrule
BNN \cite{bnn}   & .04 $\pm$ .01 &  -  & .09 $\pm$ .04 &  -  \\
CFRW \cite{tarnet}   & .04 $\pm$ .01  &  -  & .09 $\pm$ .03  &  -  \\
CEVAEs  \cite{cevae} & .02 $\pm$ .01  &   -  &  \cellcolor{gray!25} .03 $\pm$ .01  &   -  \\
GANITE  \cite{yoon2018ganite} &  \cellcolor{gray!25} .01 $\pm$ .01  &   -  & .06 $\pm$ .03  &   -  \\
\midrule

TARNet + CGC   & .037 $\pm$ .003 &   \multirow{2}{*}{0.945} & .087 $\pm$ .008 &   \multirow{2}{*}{1.022}   \\
TARNet    & .039 $\pm$ .004  &  & .085 $\pm$ .008  &  \\

\midrule
Dragonnet + CGC   & .054 $\pm$ .010 &  \multirow{2}{*}{1.121} & .097 $\pm$ .012 &  \multirow{2}{*}{1.147} \\
Dragonnet    & .048 $\pm$ .004 &   & .085 $\pm$ .008 &  \\
\midrule
BCAUSS + CGC    & .017 $\pm$ .006 & \multirow{2}{*}{0.288} & .075 $\pm$ .007 & \multirow{2}{*}{0.796} \\
BCAUSS       & .058 $\pm$ .001 &  & .094 $\pm$ .009 &  \\
\bottomrule
\end{tabular}
}
\caption {Train and test metrics on the Jobs benchmark. Lower is better. The ``Comp" value is the ratio between the constrained and the unconstrained model.} 
\end{table}

\vspace{-4mm}
\paragraph{Jobs results.}
Jobs is a small benchmark and the performance of the models varies greatly between runs. This characteristic is evidenced in the standard deviation values, which are in the same order as the mean in most cases. Because of that, we think the use of the benchmark rests upon its ability to test that models perform well on real data. In this sense, all models perform comparably well. The lowest $ATT$ error for train is obtained by GANITE followed by the constrained version of BCAUSS while for test, the lowest error is obtained by CEVAE followed by GANITE. In this dataset, the discovery of the causal graph is especially unreliable, and this is reflected in the comparison between constrained and unconstrained models, which obtain similar results.

\section{Discussion and Conclusion}

We have presented NN-CGC, a novel method for incorporating causal information into the estimation of heterogeneous treatment effects. Our method leverages an inductive bias that reduces the error caused by spurious variable interactions and can be applied on top of other representation-based models. We have tested the effectiveness of our method using three different base models. The experiments indicate that constraining models using the described method leads to significant improvements, achieving new state-of-the-art results. 

NN-CGC is flexible and performed well for all three base models. Following the idea of stacking mechanism to address sources of error, NN-CGC can be combined with other techniques to introduce additional inductive biases. We have also shown that our method is robust to imperfect causal graphs, and that using partial causal information is preferable to ignoring it. Having no restrictions on interactions is the same as assuming a causal graph that is fully connected, and in most cases there are at least some known edges that are not causal. 

As a future work, we plan to enhance our method by using masking and a graphical conditioner, a neural network that prevents spurious interactions by design, to share some weights between the groups. This would reduce redundancy in the architecture and make a more efficient use of the data.

\bibliographystyle{named}
\bibliography{ijcai24}

\begin{thebibliography}{}

\bibitem[\protect\citeauthoryear{Bengio \bgroup \em et al.\egroup }{2013}]{bengio2013representation}
Yoshua Bengio, Aaron Courville, and Pascal Vincent.
\newblock Representation learning: A review and new perspectives.
\newblock {\em IEEE transactions on pattern analysis and machine intelligence}, 35(8):1798--1828, 2013.

\bibitem[\protect\citeauthoryear{Chatton and Rohrer}{2023}]{chatton2023causal}
Arthur Chatton and Julia~M Rohrer.
\newblock The causal cookbook: Recipes for propensity scores, g-computation, and doubly robust standardization.
\newblock 2023.

\bibitem[\protect\citeauthoryear{Chen \bgroup \em et al.\egroup }{2020}]{chen2020causalml}
Huigang Chen, Totte Harinen, Jeong-Yoon Lee, Mike Yung, and Zhenyu Zhao.
\newblock Causalml: Python package for causal machine learning.
\newblock {\em arXiv preprint arXiv:2002.11631}, 2020.

\bibitem[\protect\citeauthoryear{Cinelli \bgroup \em et al.\egroup }{2022}]{good_controls}
Carlos Cinelli, Andrew Forney, and Judea Pearl.
\newblock A crash course in good and bad controls.
\newblock {\em Sociological Methods \& Research}, page 00491241221099552, 2022.

\bibitem[\protect\citeauthoryear{Curth and van~der Schaar}{2021}]{snet}
Alicia Curth and Mihaela van~der Schaar.
\newblock Nonparametric estimation of heterogeneous treatment effects: From theory to learning algorithms.
\newblock In {\em International Conference on Artificial Intelligence and Statistics}, pages 1810--1818. PMLR, 2021.

\bibitem[\protect\citeauthoryear{Curth \bgroup \em et al.\egroup }{2021}]{curthbenchmark}
Alicia Curth, David Svensson, Jim Weatherall, and Mihaela van~der Schaar.
\newblock Really doing great at estimating cate? a critical look at ml benchmarking practices in treatment effect estimation.
\newblock In {\em Thirty-fifth conference on neural information processing systems datasets and benchmarks track (round 2)}, 2021.

\bibitem[\protect\citeauthoryear{Dehejia and Wahba}{2002}]{lalondefeatureset}
Rajeev~H Dehejia and Sadek Wahba.
\newblock Propensity score-matching methods for nonexperimental causal studies.
\newblock {\em Review of Economics and statistics}, 84(1):151--161, 2002.

\bibitem[\protect\citeauthoryear{Dorie}{2016}]{dorie2016npci}
Vincent Dorie.
\newblock Npci: Non-parametrics for causal inference.
\newblock {\em URL: https://github. com/vdorie/npci}, 11:23, 2016.

\bibitem[\protect\citeauthoryear{Doutreligne and Varoquaux}{2023}]{selectcausal}
Matthieu Doutreligne and Ga{\"e}l Varoquaux.
\newblock How to select predictive models for decision making or causal inference.
\newblock {\em Available at SSRN 4467871}, 2023.

\bibitem[\protect\citeauthoryear{Harring \bgroup \em et al.\egroup }{2015}]{spurious_linear}
Jeffrey~R Harring, Brandi~A Weiss, and Ming Li.
\newblock Assessing spurious interaction effects in structural equation modeling: A cautionary note.
\newblock {\em Educational and psychological measurement}, 75(5):721--738, 2015.

\bibitem[\protect\citeauthoryear{Hassanpour and Greiner}{2019}]{hassanpour2019counterfactual}
Negar Hassanpour and Russell Greiner.
\newblock Counterfactual regression with importance sampling weights.
\newblock In {\em IJCAI}, pages 5880--5887, 2019.

\bibitem[\protect\citeauthoryear{Henckel \bgroup \em et al.\egroup }{2022}]{henckel2022graphical}
Leonard Henckel, Emilija Perkovi{\'c}, and Marloes~H Maathuis.
\newblock Graphical criteria for efficient total effect estimation via adjustment in causal linear models.
\newblock {\em Journal of the Royal Statistical Society Series B: Statistical Methodology}, 84(2):579--599, 2022.

\bibitem[\protect\citeauthoryear{Hill}{2011}]{IHDP}
Jennifer~L Hill.
\newblock Bayesian nonparametric modeling for causal inference.
\newblock {\em Journal of Computational and Graphical Statistics}, 20(1):217--240, 2011.

\bibitem[\protect\citeauthoryear{Johansson \bgroup \em et al.\egroup }{2016}]{bnn}
Fredrik Johansson, Uri Shalit, and David Sontag.
\newblock Learning representations for counterfactual inference.
\newblock In {\em International conference on machine learning}, pages 3020--3029. PMLR, 2016.

\bibitem[\protect\citeauthoryear{Kaddour \bgroup \em et al.\egroup }{2022}]{causalml_survey}
Jean Kaddour, Aengus Lynch, Qi~Liu, Matt~J Kusner, and Ricardo Silva.
\newblock Causal machine learning: A survey and open problems.
\newblock {\em arXiv preprint arXiv:2206.15475}, 2022.

\bibitem[\protect\citeauthoryear{Koller and Friedman}{2009}]{causal_graph}
Daphne Koller and Nir Friedman.
\newblock {\em Probabilistic graphical models: principles and techniques}.
\newblock MIT press, 2009.

\bibitem[\protect\citeauthoryear{K{\"u}nzel \bgroup \em et al.\egroup }{2019}]{kunzel2019metalearners}
S{\"o}ren~R K{\"u}nzel, Jasjeet~S Sekhon, Peter~J Bickel, and Bin Yu.
\newblock Metalearners for estimating heterogeneous treatment effects using machine learning.
\newblock {\em Proceedings of the national academy of sciences}, 116(10):4156--4165, 2019.

\bibitem[\protect\citeauthoryear{LaLonde}{1986}]{lalonde1986evaluating}
Robert~J LaLonde.
\newblock Evaluating the econometric evaluations of training programs with experimental data.
\newblock {\em The American economic review}, pages 604--620, 1986.

\bibitem[\protect\citeauthoryear{Lin \bgroup \em et al.\egroup }{2023}]{lin2023estimating}
Xiaofeng Lin, Guoxi Zhang, Xiaotian Lu, Han Bao, Koh Takeuchi, and Hisashi Kashima.
\newblock Estimating treatment effects under heterogeneous interference.
\newblock In {\em Joint European Conference on Machine Learning and Knowledge Discovery in Databases}, pages 576--592. Springer, 2023.

\bibitem[\protect\citeauthoryear{Louizos \bgroup \em et al.\egroup }{2017}]{cevae}
Christos Louizos, Uri Shalit, Joris~M Mooij, David Sontag, Richard Zemel, and Max Welling.
\newblock Causal effect inference with deep latent-variable models.
\newblock {\em Advances in neural information processing systems}, 30, 2017.

\bibitem[\protect\citeauthoryear{Maclaren and Nicholson}{2019}]{estimability}
Oliver~J Maclaren and Ruanui Nicholson.
\newblock What can be estimated? identifiability, estimability, causal inference and ill-posed inverse problems.
\newblock {\em arXiv preprint arXiv:1904.02826}, 2019.

\bibitem[\protect\citeauthoryear{Neal}{2020}]{neal2020introduction}
Brady Neal.
\newblock Introduction to causal inference.
\newblock {\em Course Lecture Notes (draft)}, 2020.

\bibitem[\protect\citeauthoryear{Nie and Wager}{2021}]{rlearner}
Xinkun Nie and Stefan Wager.
\newblock Quasi-oracle estimation of heterogeneous treatment effects.
\newblock {\em Biometrika}, 108(2):299--319, 2021.

\bibitem[\protect\citeauthoryear{Oprescu \bgroup \em et al.\egroup }{2023}]{blearner}
Miruna Oprescu, Jacob Dorn, Marah Ghoummaid, Andrew Jesson, Nathan Kallus, and Uri Shalit.
\newblock B-learner: Quasi-oracle bounds on heterogeneous causal effects under hidden confounding.
\newblock {\em arXiv preprint arXiv:2304.10577}, 2023.

\bibitem[\protect\citeauthoryear{Parafita and Vitri{\`a}}{2022}]{parafita2022estimand}
{\'A}lvaro Parafita and Jordi Vitri{\`a}.
\newblock Estimand-agnostic causal query estimation with deep causal graphs.
\newblock {\em IEEE Access}, 10:71370--71386, 2022.

\bibitem[\protect\citeauthoryear{Pearl}{1993}]{backdoor_criterion}
Judea Pearl.
\newblock Bayesian analysis in expert systems: comment: graphical models, causality and intervention.
\newblock {\em Statistical Science}, 8(3):266--269, 1993.

\bibitem[\protect\citeauthoryear{Pearl}{2009}]{pearl2009causality}
Judea Pearl.
\newblock {\em Causality}.
\newblock Cambridge university press, 2009.

\bibitem[\protect\citeauthoryear{Rotnitzky and Smucler}{2020}]{smucler}
Andrea Rotnitzky and Ezequiel Smucler.
\newblock Efficient adjustment sets for population average causal treatment effect estimation in graphical models.
\newblock {\em The Journal of Machine Learning Research}, 21(1):7642--7727, 2020.

\bibitem[\protect\citeauthoryear{Rubin}{1974}]{rubin1974estimating}
Donald~B Rubin.
\newblock Estimating causal effects of treatments in randomized and nonrandomized studies.
\newblock {\em Journal of educational Psychology}, 66(5):688, 1974.

\bibitem[\protect\citeauthoryear{Sch{\"o}lkopf \bgroup \em et al.\egroup }{2021}]{scholkopf2021toward}
Bernhard Sch{\"o}lkopf, Francesco Locatello, Stefan Bauer, Nan~Rosemary Ke, Nal Kalchbrenner, Anirudh Goyal, and Yoshua Bengio.
\newblock Toward causal representation learning.
\newblock {\em Proceedings of the IEEE}, 109(5):612--634, 2021.

\bibitem[\protect\citeauthoryear{Shalit \bgroup \em et al.\egroup }{2017}]{tarnet}
Uri Shalit, Fredrik~D. Johansson, and David Sontag.
\newblock Estimating individual treatment effect: generalization bounds and algorithms.
\newblock In Doina Precup and Yee~Whye Teh, editors, {\em Proceedings of the 34th International Conference on Machine Learning}, volume~70 of {\em Proceedings of Machine Learning Research}, pages 3076--3085. PMLR, 06--11 Aug 2017.

\bibitem[\protect\citeauthoryear{Shi \bgroup \em et al.\egroup }{2019}]{dragonnet}
Claudia Shi, David Blei, and Victor Veitch.
\newblock Adapting neural networks for the estimation of treatment effects.
\newblock In H.~Wallach, H.~Larochelle, A.~Beygelzimer, F.~d\textquotesingle Alch\'{e}-Buc, E.~Fox, and R.~Garnett, editors, {\em Advances in Neural Information Processing Systems}, volume~32. Curran Associates, Inc., 2019.

\bibitem[\protect\citeauthoryear{Shimizu \bgroup \em et al.\egroup }{2006}]{icalingam}
Shohei Shimizu, Patrik~O Hoyer, Aapo Hyv{\"a}rinen, Antti Kerminen, and Michael Jordan.
\newblock A linear non-gaussian acyclic model for causal discovery.
\newblock {\em Journal of Machine Learning Research}, 7(10), 2006.

\bibitem[\protect\citeauthoryear{Tesei \bgroup \em et al.\egroup }{2023}]{bcauss}
Gino Tesei, Stefanos Giampanis, Jingpu Shi, and Beau Norgeot.
\newblock Learning end-to-end patient representations through self-supervised covariate balancing for causal treatment effect estimation.
\newblock {\em Journal of Biomedical Informatics}, 140:104339, 2023.

\bibitem[\protect\citeauthoryear{Thoresen}{2019}]{thoresen2019spurious}
Magne Thoresen.
\newblock Spurious interaction as a result of categorization.
\newblock {\em BMC medical research methodology}, 19(1):1--8, 2019.

\bibitem[\protect\citeauthoryear{Xia \bgroup \em et al.\egroup }{2021}]{xia2021causal}
Kevin Xia, Kai-Zhan Lee, Yoshua Bengio, and Elias Bareinboim.
\newblock The causal-neural connection: Expressiveness, learnability, and inference.
\newblock {\em Advances in Neural Information Processing Systems}, 34:10823--10836, 2021.

\bibitem[\protect\citeauthoryear{Yoon \bgroup \em et al.\egroup }{2018}]{yoon2018ganite}
Jinsung Yoon, James Jordon, and Mihaela Van Der~Schaar.
\newblock Ganite: Estimation of individualized treatment effects using generative adversarial nets.
\newblock In {\em International conference on learning representations}, 2018.

\bibitem[\protect\citeauthoryear{Zhang \bgroup \em et al.\egroup }{2021}]{zhang2021gcastle}
Keli Zhang, Shengyu Zhu, Marcus Kalander, Ignavier Ng, Junjian Ye, Zhitang Chen, and Lujia Pan.
\newblock gcastle: A python toolbox for causal discovery.
\newblock {\em arXiv preprint arXiv:2111.15155}, 2021.

\end{thebibliography}

\end{document}